\documentclass{article}
\usepackage{ijcai24}

\usepackage{times}
\usepackage{soul}
\usepackage{url}
\usepackage[hidelinks]{hyperref}
\usepackage[utf8]{inputenc}
\usepackage[small]{caption}
\usepackage{graphicx}
\usepackage{amsmath}
\usepackage{amsthm}
\usepackage{booktabs}
\usepackage{algorithm}
\usepackage{algorithmic}
\usepackage[switch]{lineno}
\usepackage{natbib}

\usepackage{bbm}
\usepackage{dsfont}
\usepackage{color}
\usepackage{xcolor}
\usepackage{multirow}
\usepackage{pifont}

\usepackage[toc,page]{appendix}
\usepackage{amsfonts}
\usepackage{dsfont}
\usepackage{amssymb}
\usepackage{makecell}

\title{VCC-INFUSE: Towards Accurate and Efficient Selection of Unlabeled Examples in Semi-supervised Learning}

\author{
    Author Name
    \affiliations
    Affiliation
    \emails
    email@example.com
}

\author{
Shijie Fang$^{1,2}$\footnote{Work done during study at Peking University.}\footnote{Equal Contribution.}\and
Qianhan Feng$^{1\dagger}$\and
Tong Lin$^{1}$\footnote{Corresponding Author.}\\
\affiliations
$^1$National Key Laboratory of General Artificial Intelligence, \\School of Intelligence Science and Technology, Peking University\\
$^2$Google, Shanghai, China\\
\emails
shijiefang@google.com,
fengqianhan@stu.pku.edu.cn,
lintong@pku.edu.cn
}

\begin{document}

\maketitle

\begin{abstract}
    Despite the progress of Semi-supervised Learning (SSL), existing methods fail to utilize unlabeled data effectively and efficiently. Many pseudo-label-based methods select unlabeled examples based on inaccurate confidence scores from the classifier. Most prior work also uses all available unlabeled data without pruning, making it difficult to handle large amounts of unlabeled data. To address these issues, we propose two methods: Variational Confidence Calibration (\textbf{VCC}) and Influence-Function-based Unlabeled Sample Elimination (\textbf{INFUSE}). VCC is a universal plugin for SSL confidence calibration, using a variational autoencoder to select more accurate pseudo labels based on three types of consistency scores. INFUSE is a data pruning method that constructs a core dataset of unlabeled examples under SSL. Our methods are effective in multiple datasets and settings, reducing classification error rates and saving training time. 
    Together, VCC-INFUSE reduces the error rate of FlexMatch on the CIFAR-100 dataset by 1.08\% while saving nearly half of the training time.
\end{abstract}

\section{Introduction}
Deep neural networks underpin various machine learning applications, with their success attributed in part to extensive labeled datasets like ImageNet \citep{DBLP:conf/cvpr/DengDSLL009} and COCO \citep{DBLP:conf/eccv/LinMBHPRDZ14}. However, the process of collecting and annotating large datasets is resource-intensive and raises privacy concerns, making the acquisition of unlabeled data a more feasible and cost-effective alternative.

To address the challenge of limited labeled examples, semi-supervised learning (SSL) has gained prominence for leveraging abundant unlabeled data. Pseudo-labeling is a common SSL approach, as demonstrated by FixMatch \citep{DBLP:conf/nips/SohnBCZZRCKL20}. FixMatch generates pseudo labels for unlabeled data based on model predictions. The threshold-based selection module in FixMatch filters examples with confidence scores surpassing a fixed threshold $\tau$ for training. Formally, the loss on unlabeled data is defined as:

\begin{equation}
\label{eq:fixmatch}
\mathcal{L}_{unlab} = \sum_{i} \mathbbm{1} (\max (c_i) \ge \tau) \mathcal{L} (\hat c_i, \widetilde c_i),
\end{equation}
where $\mathcal{L} (\hat c_i, \widetilde c_i)$ represents the loss between class label and confidence distribution.

Despite FixMatch's wide adoption, it may encounter challenges in utilizing unlabeled examples effectively and efficiently. Specifically, (1) \textbf{Incorrect pseudo labels} may arise due to calibration errors in model predictions, leading to unreliable performance. (2) The \textbf{significant computation cost} involved in forwarding the entire unlabeled dataset for pseudo label selection may be alleviated by dynamically pruning the dataset. This ensures that only informative data points contribute to the model's decision boundary, reducing computation overhead and accelerating convergence.

In this paper, we propose solutions to address these challenges, enhancing the reliability and efficiency of SSL methods based on pseudo-labeling. To address the first issue, we introduce Variational Confidence Calibration (VCC), a method aimed at obtaining calibrated confidence scores for pseudo label selection. The calibrated confidence score, closely aligned with the ground-truth probability of correct predictions, serves as a more reliable metric for choosing pseudo-labeled examples. While confidence calibration is a well-explored concept in fully-supervised settings, its application in semi-supervised learning (SSL) is more challenging due to the absence of ground-truth labels. To overcome this challenge, we utilize three consistency scores to assess prediction stability. By simultaneously considering both stability and confidence, we approximate calibrated confidence scores. Additionally, a variational autoencoder enhances stability by reconstructing the calibrated confidences.

To address the second issue, we propose INFUSE (Influence Function-based Unlabeled Sample Elimination), a method leveraging influence functions \citep{DBLP:conf/icml/KohL17} to compute the importance of each unlabeled example. INFUSE dynamically retains data points with the highest importance, forming a smaller core set to replace the entire dataset. This core set allows for faster model convergence, reducing computation costs during training. The combined VCC-INFUSE method enhances prediction accuracy while minimizing training costs.

In summary, this paper makes following contributions:
\begin{itemize}
    \item We propose the VCC method, which generates well-calibrated confidence scores for more accurate pseudo labels, enhancing model accuracy. As a flexible, plug-and-play module, VCC can be seamlessly integrated with existing SSL methods.
    \item We introduce the INFUSE method, which dynamically prunes unimportant unlabeled examples to expedite convergence and reduce computation costs during training.
    \item The effectiveness of our methods is demonstrated across multiple datasets and various settings.
\end{itemize}

\section{Related Work}
\textbf{Semi-Supervised Learning.} FixMatch \citep{DBLP:conf/nips/SohnBCZZRCKL20} stands out as one of the most widely adopted SSL methods. FixMatch utilizes a weakly-augmented unlabeled example to obtain a one-hot pseudo label, followed by training the model on strongly-augmented examples to produce predictions consistent with the pseudo label. FlexMatch \citep{DBLP:conf/nips/ZhangWHWWOS21} introduces an adaptive threshold strategy, tailored to different learning stages and categories. SimMatch \citep{DBLP:conf/cvpr/ZhengYHWQX22} considers both semantic and instance similarity, promoting consistent predictions and similar similarity relationships for the same instance. Additionally, explicit consistency regularization is employed in various SSL methods \citep{DBLP:journals/corr/LaineA16,DBLP:conf/iclr/BerthelotCCKSZR20,DBLP:journals/pami/MiyatoMKI19,ganev2020semi,DBLP:conf/iclr/0102TFW0S0RS23,DBLP:conf/iccv/0001XH21,DBLP:conf/aaai/FengXF024,lee2021abc}.

\textbf{Confidence Calibration.} \cite{DBLP:conf/icml/GuoPSW17} identified the calibration problem in modern classifiers and proposed Temperature Scaling (TS) to rescale confidence distributions, preventing over-confidence. Ensemble TS \citep{DBLP:conf/icml/ZhangKH20} extends TS's representation ability by expanding the parameter space. Additionally, \citet{DBLP:conf/icml/KumarSJ18} introduces the MMCE method, a trainable calibration regularization based on Reproducing Kernel Hilbert Space (RKHS). Notably, these methods are designed for fully-supervised settings where ground-truth labels are available.

\textbf{Core Set Selection.} While most core set selection methods focus on the fully-supervised setting, our work aligns more closely with the semi-supervised learning context. \citet{DBLP:conf/nips/PaulGD21} proposes the EL2N method, measuring the importance of an example based on the norm of the loss. EL2N significantly reduces training time with a minimal impact on accuracy. GradMatch \citep{DBLP:conf/icml/KillamsettySRDI21} extends the core dataset to a weighted set using a submodular function. In the realm of SSL, RETRIEVE \citep{DBLP:conf/nips/KillamsettyZCI21} addresses core set selection as an optimization problem. However, RETRIEVE's optimization function only considers the loss labeled set, potentially deviating from the desired objective of minimizing loss on the validation set.

\begin{figure}[t]
\centering
\includegraphics[width=3.2cm]{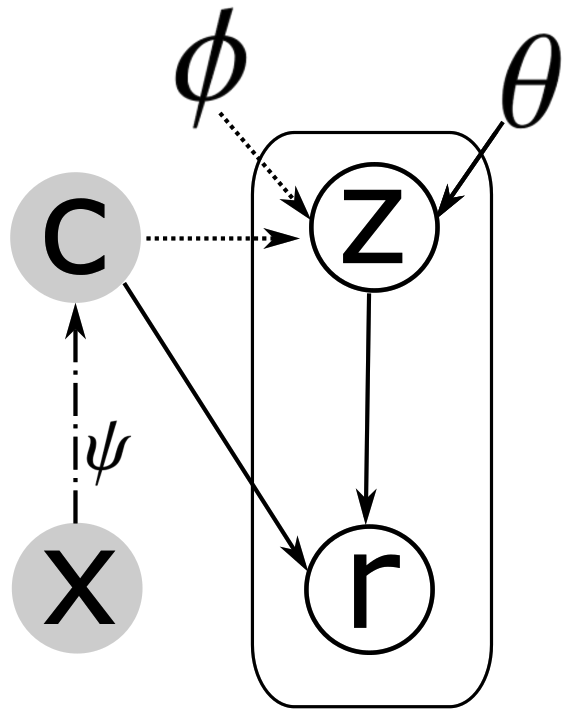}
\caption{The graphical model of VCC. Here $x$ is the input, $c$ is the originally predicted confidence distribution, $z$ is the latent variable sampled from the encoder, and $r$ is the reconstructed confidence for pseudo label selection. Dash-dotted line denotes the original prediction function $p_\psi(c | x)$.
Solid lines denote the generative model $p_\theta(r|c,z,x)$. Dashed lines denote the approximation $q_\phi(z|c,x)$ to the intractable posterior $p_\theta(z|c,r,x)$. The variational parameter $\phi$ is learned jointly with the generative model parameter $\theta$.}
\label{vcc_prob}
\end{figure}

\section{Confidence Calibration with VCC}
Many existing calibration methods are ill-suited for SSL due to the absence of ground-truth labels for unlabeled examples. Directly using the original confidence score for pseudo label selection can yield noisy results. To tackle this challenge, we introduce three different consistency scores ($s^{ens}$, $s^{tem}$, and $s^{view}$) to simultaneously gauge the stability of predictions. By combining these three scores, we obtain the approximated calibrated confidence $\tilde{r}$, which is closer to the probability of an example being correctly classified. However, $\tilde{r}$ is not directly utilized for pseudo label selection, as the process of estimating $\tilde{r}$ from three consistency scores can still be unstable for some examples.

\begin{figure*}[t] 
\centering
\includegraphics[width=0.8\textwidth]{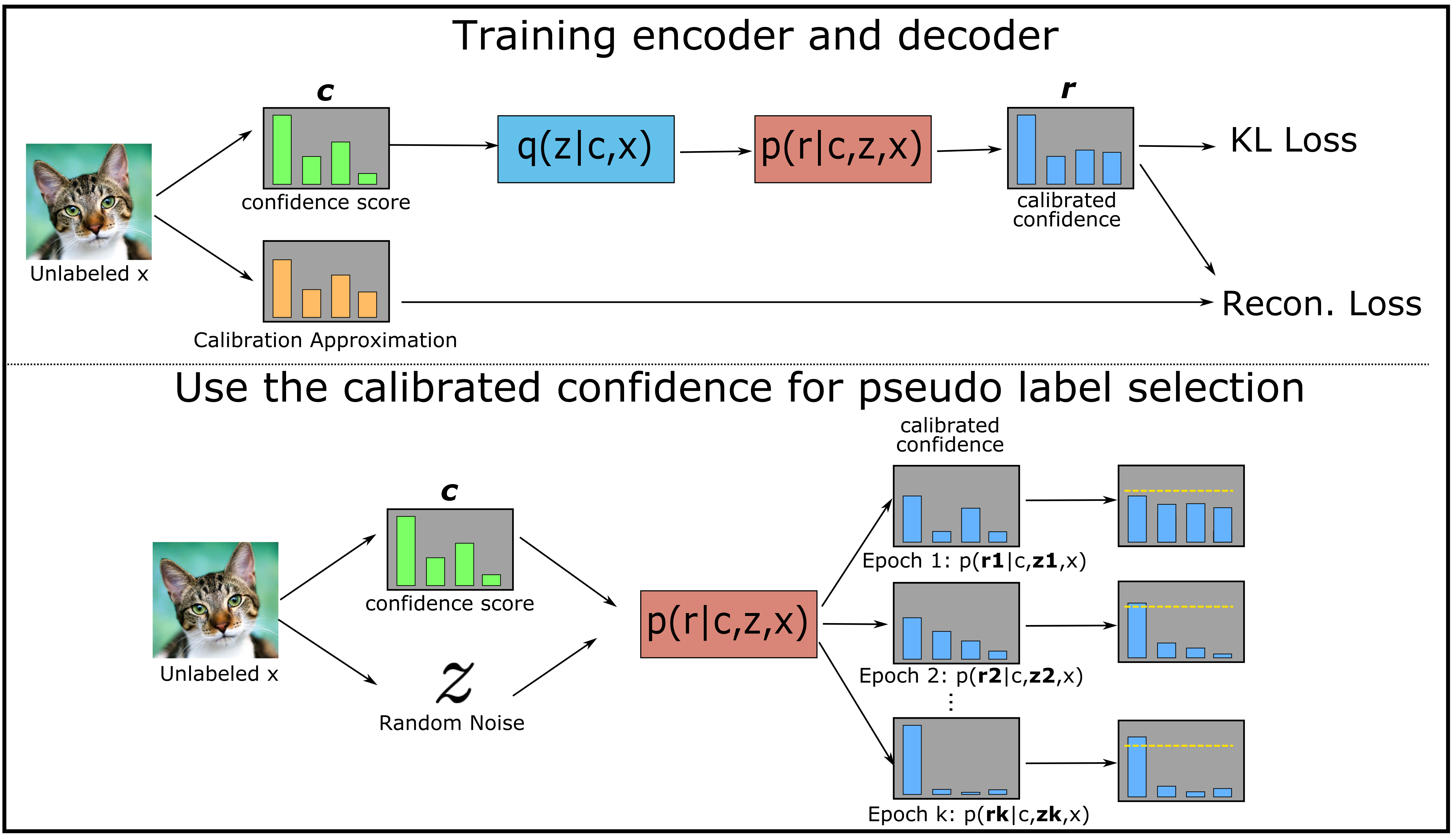}
\caption{Illustration of VCC: training VAE and using the reconstructed confidence for pseudo label selection.}
\label{vcc_framework}
\end{figure*}

To mitigate this instability, we introduce a Variational Autoencoder (VAE) to reconstruct $\tilde{r}$ for pseudo label selection. The graphical model and framework illustration of VCC are provided in Fig.\ref{vcc_prob} and \ref{vcc_framework}, respectively. The VAE is learned jointly with the original classifier during training, where $\tilde{r}$ serves as the "ground-truth" for calculating the reconstruction loss. For pseudo label selection, we leverage the output of the VAE as the calibrated confidence.

\subsection{Ensemble Consistency}
From a Bayesian perspective, the parameters $\theta$ of a model are sampled from a probability distribution over the training set $D$. Model's prediction for sample $x$ can be formulated as:
\begin{align}
p(y | x, D)=\int p(y | x, \theta) p(\theta | D) \mathrm{d} \theta,
\label{c03_eq_bayesian_int}
\end{align}
where $p(y | x, \theta)$ represents the probability distribution of the label $y$ of $x$ given the parameters $\theta$, and $p(\theta | D)$ represents the probability distribution of the model parameters $\theta$ trained on the dataset $D$.
A single model may provide incorrect predictions, for example, $x$ due to randomness and noise, even if the confidence is high. Considering the entire parameter space, if all model parameters yield consistent predictions for $x$, the result is more convincing. In this case, the prediction can be viewed as an ensemble of predictions from multiple models.

However, due to the large parameter space of $\theta$, direct computation of Equation \ref{c03_eq_bayesian_int} is intractable. Therefore, we apply Monte-Carlo Dropout \citep{DBLP:conf/icml/GalG16} on the linear layer to approximate the computation of Equation \ref{c03_eq_bayesian_int}. The feature map is cloned by $K$ copies, followed by a Dropout layer to randomly eliminate neural connections in the classification head to obtain predictions. By doing so, the model would generate $K$ estimated confidence distributions of example $i$, and the expectation can be treated as the ensemble of $K$ different models:
\begin{equation}
\hat{\mathbf{y}_i} = p(y | x, \text{Dropout}(\theta)), \quad
\tilde{\mathbf{y}} = \frac{1}{K} \sum_{i=1}^K \hat{\mathbf{y}}_{i}.
\end{equation}
Then, entropy is employed as the ensemble-consistency score to measure the different models' consistency of example: $s^{ens} = -\sum_{c=1}^M \tilde{\mathbf{y}}_c \log \tilde{\mathbf{y}}_c$.

\subsection{Temporal Consistency}
In SSL, parameters are updated frequently during training, making the decision boundaries change all the time. Some examples may shift from one side of the decision boundary to the other side after parameter updates, bringing a change in classification results. In this case, the prediction results of many examples may be rather unstable. If these examples are used in training, it may result in incorrect pseudo labels and hinder the model's performance.

To measure the stability of prediction results between different steps, we propose the temporal consistency score, which considers the changes in the confidence distribution of an example between different epochs. Specifically, let $y^t$ represent the confidence distribution of an example at epoch $t$. The temporal consistency score can be calculated as:
\begin{align}
\begin{split}
    s^{tem} 
&= D_{KL}\left(y^t \Bigg{\Vert} \frac{1}{K} \sum_{k=1}^K y^{t-k}\right) \label{c03_eq_temporal_score} \\
&= \sum_{c=1}^M y^t_c \log \left( \frac{y^t_c}{\frac{1}{K} \sum_{k=1}^K y^{t-k}_c} \right),
\end{split}
\end{align}
where $D_{KL}$ represents the Kullback-Leibler Divergence, $M$ is the number of classes, $K$ represents the window size. In experiments, we empirically set $K=1$ to preserve the sensitivity of abnormal confidences. Although both consider the problem from the perspective of time, our method differs a lot from the method proposed by \cite{DBLP:conf/icml/ZhouWB20}.

\subsection{View Consistency}
Multi-view learning \citep{DBLP:journals/tip/XuT015} aims to leverage multiple perspectives to predict data, allowing different predictors to correct predictions collectively.

In semi-supervised learning (SSL), obtaining models with different views often involves dividing the entire dataset into multiple subsets for training multiple models. However, this incurs high model training costs, and the volume of labeled data in each subset may be too small to train a decent model. To address this, we use Exponential Moving Average (EMA) to construct models with different views. The original model parameter $\theta$ is updated using gradient descent, while $\theta_{ema}$ is updated using the EMA scheme:
\begin{align}
\theta^t_{ema} = \theta^t \cdot \beta + \theta^{t-1}_{ema} \cdot (1 - \beta),
\label{ema_update}
\end{align}
where $\beta$ is a decay hyperparameter. These can be treated as two different views from the same network structure.

A typical classification model consists of a feature extraction network (backbone) and a classification head (linear layer). To increase the difference between two views, we adopt a cross-feature trick. The backbone of each view first extracts features from input, which are then fed into the classification head of another view. This can be formulated as:
\begin{align}
y = p(y | x, \theta^{backbone}, \theta^{head}_{ema}), \\
\quad
y_{ema} = p(y | x, \theta^{backbone}_{ema}, \theta^{head}).
\end{align}
After obtaining the outputs, the Kullback-Leibler (KL) divergence is used to measure the consistency between them:
\begin{align}
    s^{view} &= D_{KL} \left(y || y_{ema} \right). \label{c03_eq_view_score}
\end{align}

It may seem like temporal consistency and view consistency overlap to some extent, as the predictions of the EMA model used in view consistency can also be considered an ensemble of predictions from past epochs. The difference is that the cross-feature trick is used in the computation of view consistency, which enforces this metric to focus more on consistency over multiple views rather than multiple time steps.

\subsection{Approximation of Calibrated Confidence}
We have introduced three scores to evaluate the stability of predictions. However, $s^{ens}$, $s^{tem}$, and $s^{view}$ cannot be directly used for pseudo label selection, which is based on confidence scores. To address this, we propose a simple method to approximate calibrated confidence with the three consistency scores.  


The consistency scores are first normalized and summed up as the stability score. First, a fixed-length queue $q$ is maintained to record the historical predictions of the unlabeled samples in mini-batches. Since $s^{ens}$, $s^{tem}$, and $s^{view}$ have different distributions, we normalize them with max-min normalization. Let $u$ be the unlabeled example, the normalization is done as follows:
\begin{align}
    \tilde{s}^{t}_u &= \frac{s^{t}_u - \min_{u' \in q}\left( s^{t}_{u'} \right)}{\max_{u' \in q}\left( s^{t}_{u'} \right) - \min_{u' \in q}\left( s^{t}_{u'} \right)},
\end{align}
where $t=\{ens,tem,view\}$. After normalization, $\tilde{s}^{ens}_u$, $\tilde{s}^{tem}_u$, and $\tilde{s}^{view}_u$ are all real numbers ranging from 0 to 1. These consistency scores evaluate the stability of the examples. However, some hard-to-learn examples may also have stable predictions but low confidence scores. Hence, the three consistency scores are not enough to describe the reliability of the prediction. To address this, the original confidence score of the sample $\tilde{s}^{conf}_u = \max\left( y_u \right)$ is also used. Thus, an unlabeled sample $u$ can be represented by a quadruple $(\tilde{s}^{ens}_u, \tilde{s}^{tem}_u, \tilde{s}^{view}_u, \tilde{s}^{conf}_u)$.

The next problem is how to combine these four scores together for estimation. To avoid complex parameter tuning, VCC adopts a simple yet effective approach: taking the sum of their squares:
\begin{align}
s_u = \left(\tilde{s}^{ens}_u\right)^2 + \left(\tilde{s}^{tem}_u\right)^2 + \left(\tilde{s}^{view}_u\right)^2 + \left(\tilde{s}^{conf}_u\right)^2.
\label{c03_eq_score_fusion}
\end{align}

According to the results in \cite{DBLP:conf/icml/GuoPSW17}, calibration errors mainly occur in the middle range of confidences, while samples with extremely low or high confidences tend to have smaller calibration errors. 
Therefore, we approximately treat the lowest/highest confidence score in $q$ as well-calibrated and employ interpolation to calculate the calibrated confidence scores for other examples. To further eliminate the unfairness between different categories, the interpolation operation only considers examples with the same pseudo labels as the current example $u$.
\begin{equation}
    \begin{split}
    & q'  = \{ e \mid e \in q, \arg\max \tilde{s}^{conf}_e = \arg\max \tilde{s}^{conf}_u \}, \\
    & max\_score = \max_{u' \in q'} \left( {s}_{u'} \right),
    min\_score = \min_{u' \in q'} \left( {s}_{u'} \right), \\
    & max\_conf = \max_{u' \in q'} \left( \tilde{s}_{u'}^{conf} \right),
    min\_conf = \min_{u' \in q'} \left( \tilde{s}_{u'}^{conf} \right), 
    \end{split}
    \label{c03_eq_cali_approximation}
\end{equation}
and $\tilde{r}_u$ can be formulated as: 
\begin{equation}
    \begin{split}
        \tilde{r}_u = \frac{max\_score - s_u}{max\_score- min\_score} \cdot (max\_conf &- min\_conf) \\
        &+ min\_conf. \notag
    \end{split}
\end{equation}

\label{section_calibration_approx}

\subsection{Reconstruct $\tilde{r}_u$ with Variational Autoencoder}
In Section \ref{section_calibration_approx}, we combined three consistency scores to obtain $\tilde{r}_u$, which is the approximation of calibrated confidence scores. However, it may face instability due to the update of queue $q$ and abnormal interpolation endpoints. To address this, we reconstruct the statistical-based $\tilde{r}_u$ in a learning-based way. Specifically, a Variational Autoencoder (VAE) is employed to generate the calibrated confidence score $r_u$ for pseudo label selection, and $\tilde{r}_u$ is used as input for training the VAE. 

We assume $r$ is generated by the following random process, which includes two steps: (1) a hidden variable $z$ sampled from a prior distribution $p_\theta(z)$; (2) a value $r$ generated from the conditional distribution $p_\theta(r|c,z,x)$:
\begin{align}
    \label{vc_eq1}
    p_\theta(r|c,x) = \int_z p_\theta(z) p_\theta(r|z,c,x) \mathrm{d}z.
\end{align}

However, the marginal likelihood $p_\theta(r|c,x)$ is generally intractable. Hence another distribution $q_\phi(z|c,x)$ is introduced as the approximation of $p_\theta(z)$ (Please refer to Appendix \ref{appendix_vcc_formulation} for details):
\begin{align}
\begin{split}
& \log p_\theta(r |c,x) = \int_z q_\phi(z|c,x) \log p_\theta(r|c,x) \mathrm{d}z \\
& \ge \mathbb{E}_{q_\phi (z|c,x)} \log p_\theta (r|c,z,x)- D_{KL} (q_\phi (z|c,x) \Vert p_\theta(z|c,x)).
\end{split}
\label{vc_eq2}
\end{align}

The first term is the likelihood of calibration reconstruction (denoted as $\mathcal{L}_{VCC}^{recon}$), where $q_\phi(z|c,x)$ is the encoder to infer the hidden variable $z$, and $p_\theta(r|c,z,x)$ is the decoder to recover a calibrated confidence $r$. To compute the reconstruction loss, the approximated $\tilde{r}$ is used as the ground truth. Besides, $z$ needs to be sampled from $q_\phi(z|c,x)$. Reparameterization trick \citep{DBLP:journals/corr/KingmaW13} is used to predict the mean and standard deviation of $z$. By setting $\epsilon \sim \mathcal{N}(0,1)$, the reparameterization is formulated as $z = \mu(c,x) + \epsilon \cdot \sigma(c,x)$. For the second term, under the Gaussian assumptions of the prior $p_\theta(z|c,x) \sim \mathcal{N}(0,1)$ and the approximator $q_\phi(z|c,x) \sim \mathcal{N}(\mu(c,x), \sigma^2(c,x))$, we have:
\begin{align}
\label{vc_eq4}
\begin{split}
    \mathcal{L}_{VCC}^{KL} \doteq & D_{KL} (q_\phi (z|c,x) \Vert p_\theta(z|c,x)) \\
    = & - \log {\sigma} + \frac{\mu^2 + \sigma^2}{2} - \frac{1}{2}.
\end{split}
\end{align}
The overall objective function can be formulated as:
\begin{align}
    \mathcal{L} = \mathcal{L}_{lab} + \lambda_{unlab} \cdot \mathcal{L}_{unlab} + \lambda_{VCC} \cdot \left(\mathcal{L}_{VCC}^{recon} -  \mathcal{L}_{VCC}^{KL}\right).
\end{align}

Although a more accurate confidence score is generated by combining three consistencies, it is still not as optimal as the inaccessible ground-truth. This is because there are many other “nuisance” and untraceable factors that affect the pseudo label's approach toward the ground-truth, such as the randomness of the neural networks. Under these circumstances, directly approaching the unreliable target may still degrade performance. The original VAE is proposed to learn continuous distribution features from discontinuous distributions by sampling a hidden variable. This process is suitable for suboptimal pseudo label learning because the approach of the prediction to the generated pseudo label can be viewed as the process of the prediction approaching the ground-truth. Since eliminating those nuisance factors cannot be tractable, we use VAE to simulate this process instead of the MLP.
\begin{table*}[t]
\centering
\caption{Comparison of error rate (\%) for different methods under various settings.}
\scalebox{0.85}{
\begin{tabular}{c|ccc|ccc|ccc}
\toprule
\multirow{2}{*}{\textbf{Method}} & \multicolumn{3}{c|}{\textbf{CIFAR-10}} & \multicolumn{3}{c|}{\textbf{CIFAR-100}} & \multicolumn{3}{c}{\textbf{SVHN}} \\
\cmidrule(lr){2-4} \cmidrule(lr){5-7} \cmidrule(lr){8-10}
& 40 & 250 & 2500 & 400 & 2500 & 10000 & 40 & 250 & 1000 \\
\midrule
PL & $76.29_{\pm1.08}$ & $48.28_{\pm2.01}$ & $14.90_{\pm0.20}$ & $87.15_{\pm0.47}$ & $59.09_{\pm0.61}$ & $38.86_{\pm0.09}$ & $75.95_{\pm3.39}$ & $16.60_{\pm1.13}$ & $9.33_{\pm0.58}$ \\
UDA & $8.01_{\pm1.34}$ & $5.12_{\pm0.15}$ & $4.32_{\pm0.07}$ & $53.44_{\pm2.06}$ & $34.37_{\pm0.28}$ & $27.52_{\pm0.10}$ & $2.03_{\pm0.02}$ & $2.03_{\pm0.03}$ & $1.96_{\pm0.01}$ \\
VAT & $76.42_{\pm2.57}$ & $42.58_{\pm6.67}$ & $10.97_{\pm0.19}$ & $83.11_{\pm0.27}$ & $53.17_{\pm0.57}$ & $36.58_{\pm0.21}$ & $77.00_{\pm6.59}$ & $4.59_{\pm0.13}$ & $4.09_{\pm0.21}$ \\
MeanTeacher & $76.93_{\pm2.29}$ & $56.06_{\pm2.03}$ & $15.47_{\pm0.43}$ & $90.34_{\pm0.65}$ & $61.13_{\pm0.57}$ & $39.05_{\pm0.12}$ & $81.94_{\pm1.33}$ & $25.10_{\pm3.17}$ & $12.29_{\pm0.45}$ \\
MixMatch & $70.67_{\pm1.25}$ & $37.28_{\pm0.61}$ & $7.38_{\pm0.06}$ & $79.95_{\pm0.29}$ & $49.58_{\pm0.62}$ & $32.10_{\pm0.13}$ & $79.63_{\pm5.78}$ & $3.71_{\pm0.20}$ & $3.12_{\pm0.09}$ \\
ReMixMatch & $14.50_{\pm2.58}$ & $9.21_{\pm0.55}$ & $4.89_{\pm0.05}$ & $57.10_{\pm0.01}$ & $34.77_{\pm0.32}$ & $26.18_{\pm0.23}$ & $31.27_{\pm18.79}$ & $6.38_{\pm1.09}$ & $5.34_{\pm0.45}$ \\
Dash(RandAug) & $15.01_{\pm3.70}$ & $5.13_{\pm0.26}$ & $4.35_{\pm0.09}$ & $53.98_{\pm2.31}$ & $34.47_{\pm0.12}$ & $27.72_{\pm0.03}$ & $2.08_{\pm0.09}$ & $1.97_{\pm0.01}$ & $2.03_{\pm0.03}$ \\
SoftMatch & $5.06_{\pm0.02}$ & $4.84_{\pm0.10}$ & $4.27_{\pm0.12}$ & $49.64_{\pm1.46}$ & $33.05_{\pm0.05}$ & $27.26_{\pm0.03}$ & $2.31_{\pm0.01}$ & $2.15_{\pm0.05}$ & $2.08_{\pm0.04}$ \\
CoMatch & $5.44_{\pm0.05}$ & $5.33_{\pm0.12}$ & $4.29_{\pm0.04}$ & $60.98_{\pm0.77}$ & $37.24_{\pm0.24}$ & $28.15_{\pm0.16}$ & $9.51_{\pm5.59}$ & $2.21_{\pm0.20}$ & $1.96_{\pm0.07}$ \\
\hline
FixMatch & $7.52_{\pm0.42}$ & $4.90_{\pm0.03}$ & $4.28_{\pm0.10}$ & $46.47_{\pm0.05}$ & $28.09_{\pm0.06}$ & $22.21_{\pm0.02}$ & $\textbf{2.96}_{\pm1.23}$ & $1.99_{\pm0.05}$ & $1.96_{\pm0.06}$ \\
VCC-FixMatch & $\textbf{6.84}_{\pm0.52}$ & $\textbf{4.68}_{\pm0.04}$ & $\textbf{4.27}_{\pm0.21}$ & $\textbf{43.31}_{\pm0.02}$ & $\textbf{27.76}_{\pm0.06}$ & $\textbf{22.05}_{\pm0.03}$ & $3.12_{\pm0.61}$ & $\textbf{1.97}_{\pm0.02}$ & $\textbf{1.95}_{\pm0.08}$ \\
\hline
FlexMatch & $4.98_{\pm0.01}$ & $5.00_{\pm0.05}$ & $4.24_{\pm0.07}$ & $40.43_{\pm0.63}$ & $26.38_{\pm0.17}$ & $21.83_{\pm0.08}$ & $3.36_{\pm0.37}$ & $5.02_{\pm1.20}$ & $5.43_{\pm0.46}$ \\
VCC-FlexMatch & $\textbf{4.90}_{\pm0.10}$ & $\textbf{4.65}_{\pm0.07}$ & $\textbf{4.14}_{\pm0.15}$ & $\textbf{37.98}_{\pm0.65}$ & $\textbf{25.75}_{\pm0.11}$ & $\textbf{21.48}_{\pm0.07}$ & $\textbf{2.62}_{\pm0.08}$ & $\textbf{4.97}_{\pm0.08}$ & $\textbf{3.71}_{\pm1.13}$ \\
\hline
SimMatch & $5.60_{\pm1.37}$ & $4.84_{\pm0.39}$ & $3.96_{\pm0.01}$ & $37.81_{\pm2.21}$ & $25.07_{\pm0.32}$ & $\textbf{20.58}_{\pm0.11}$ & $3.70_{\pm0.72}$ & $2.27_{\pm0.12}$ & $\textbf{2.07}_{\pm0.08}$ \\
VCC-SimMatch & $\textbf{5.27}_{\pm0.34}$ & $\textbf{4.76}_{\pm0.14}$ & $\textbf{3.87}_{\pm0.24}$ & $\textbf{37.22}_{\pm0.04}$ & $\textbf{24.98}_{\pm0.13}$ & $20.61_{\pm0.01}$ & $\textbf{3.04}_{\pm0.02}$ & $\textbf{2.20}_{\pm0.01}$ & $4.39_{\pm0.02}$ \\
\hline
Fully-Supervised & \multicolumn{3}{c|}{$4.58_{\pm0.05}$} & \multicolumn{3}{c|}{$19.63_{\pm0.08}$} & \multicolumn{3}{c}{$2.07_{\pm0.02}$} \\
\bottomrule
\end{tabular}
}
\label{tab_vcc_main}%
\end{table*}%


\begin{table}[!t]
\centering
\caption{Error rate results (\%) on STL-10 dataset.}
\label{stl10_results}
\setlength{\tabcolsep}{2pt} 
\scalebox{0.85}{{%
\begin{tabular}{c|cc|cc|cc}
\toprule
Labels & FixMatch & w/ VCC & FlexMatch & w/ VCC & SimMatch & w/ VCC \\
\midrule
40 & 35.97 & \textbf{30.63} & 29.15 & \textbf{28.14} & 27.84 & \textbf{26.97} \\
1000 & 6.25 & \textbf{5.31} & 5.77 & \textbf{5.52} & 5.91 & \textbf{5.51} \\
\bottomrule
\end{tabular}%
}}
\end{table}

\begin{table*}[t]
\centering
\caption{Error rate, ECE \citep{DBLP:conf/icml/GuoPSW17}, MCE \citep{DBLP:conf/icml/GuoPSW17}, and ACE \citep{DBLP:conf/cvpr/NixonDZJT19} results on CIFAR-100 with 400/2500/10000 labeled examples.}
\label{appendix_sup_vcc_calibration_results}
\scalebox{0.8}{%
\begin{tabular}{c|cccc|cccc|cccc}
\toprule
\multirow{2}{*}{Method} & \multicolumn{4}{c|}{400 labels} & \multicolumn{4}{c|}{2500 labels} & \multicolumn{4}{c}{10000 labels} \\
\cline{2-13}
& ER(\%) & ECE & MCE & ACE & ER(\%) & ECE & MCE & ACE & ER(\%) & ECE & MCE & ACE \\
\hline
FixMatch & 46.42 & 0.382 & 0.573 & 0.376 & 28.03 & 0.208 & 0.530 & 0.199 & 22.20 & 0.127 & 0.322 & 0.128 \\
VCC-FixMatch & \textbf{43.29} & \textbf{0.359} & \textbf{0.560} & \textbf{0.345} & \textbf{27.81} & \textbf{0.195} & \textbf{0.418} & \textbf{0.182} & \textbf{22.01} & \textbf{0.125} & \textbf{0.317} & \textbf{0.127} \\
\hline
FlexMatch & 39.94 & 0.291 & 0.512 & 0.286 & 26.49 & 0.169 & 0.369 & 0.173 & 21.90 & 0.120 & 0.311 & 0.126 \\
VCC-FlexMatch & \textbf{37.52} & \textbf{0.257} & \textbf{0.446} & \textbf{0.258} & \textbf{25.26} & \textbf{0.147} & \textbf{0.324} & \textbf{0.163} & \textbf{21.55} & \textbf{0.104} & \textbf{0.269} & \textbf{0.125} \\
\hline
SimMatch & 37.81 & 0.325 & \textbf{0.510} & 0.328 & 25.07 & 0.157 & 0.358 & 0.179 & \textbf{20.58} & \textbf{0.113} & 0.295 & \textbf{0.116} \\
VCC-SimMatch & \textbf{37.20} & \textbf{0.317} & 0.514 & \textbf{0.314} & \textbf{25.01} & \textbf{0.155} & \textbf{0.347} & \textbf{0.173} & 20.61 & 0.115 & \textbf{0.291} & 0.121 \\
\bottomrule
\end{tabular}%
}
\end{table*}

\section{Core Set Selection with INFUSE}
In the previous section, we introduced VCC framework, which ensures well-calibrated confidence scores to improve accuracy in pseudo label selection. Nonetheless, as discussed earlier, training the SSL model still encounters substantial computational expenses. Furthermore, the incorporation of additional encoder and decoder of VCC introduces an extra computation overhead. To address these challenges, we present INFUSE—a core set selection methodology aimed at efficient example selection. Based on the influence function \cite{DBLP:conf/icml/KohL17}, INFUSE allows for training the SSL model using only a subset of the complete unlabeled dataset, so that training time can be significantly reduced.

In SSL, the model should minimize the loss on the validation set to obtain the highest generalization accuracy:
\begin{equation}
    \min \mathcal{L}(V, \theta^*), \quad
\text{s.t. \ } \theta^* = \mathop{\mathrm{\arg\min}}\limits_{\theta}  \  R(\theta),
\end{equation}
\begin{equation}
\begin{split}
R(\theta) \doteq &\mathbb{E}_{\left(x,y\right) \in S} \left[ H(q_x, y) \right]  \\
 & + \lambda \cdot \mathbb{E}_{u \in U} \left[ \mathds{1} \left(\max (q_u) \ge \tau \right) \cdot {H \left( \hat{q_u}, p\left(y \mid u \right)  \right) } \right]. \notag
\end{split}
\end{equation}
Here $H$ is the loss function, $\tau$ is the threshold for pseudo label selection, $q$ is the confidence distribution, $\hat{q}$ is the pseudo label, and $R(\theta)$ is the total loss on labeled dataset $S$ and unlabeled dataset $U$. Now assume the weight of an unlabeled example $u^{\prime}$ is increased by $\epsilon$. Denote $\mathcal{L}_U (u^{\prime}, \theta) = \lambda \cdot \mathds{1} \left(\max (q_{u^{\prime}}) \ge \tau \right) \cdot {H \left( \hat{q_{u^{\prime}}}, p\left(y \mid {u^{\prime}} \right)  \right) }$
, the optimal model parameters corresponding to the new training set become:
\begin{align}
    \hat{\theta} &= \mathop{\mathrm{\arg\min}}\limits_{\theta}  \  R(\theta) + \epsilon \cdot  \mathcal{L}_U (u^{\prime}, \theta).
    \label{c03_eq_infuse_eq1}
\end{align}
In Equation \ref{c03_eq_infuse_eq1}, $\hat{\theta}$ minimizes the loss function on the training set, which means the gradient w.r.t $\hat{\theta}$ is 0:
\begin{gather}
    \nabla_{\theta} R(\hat{\theta}) + \epsilon \nabla_{\theta} \mathcal{L}_U (u^{\prime}, \hat{\theta}) = 0.
    \label{c03_eq_infuse_eq2}
\end{gather}
Using a Taylor-series approximation at $\theta^*$, Equation \ref{c03_eq_infuse_eq2} can be rewritten as:
\begin{gather}
\begin{split}
    \nabla_{\theta}& R\left(\theta^{*}\right)+\epsilon \cdot \nabla_{\theta} \mathcal{L}_{U}\left(u^{\prime}, \theta^{*}\right) \\
    &+\left(\nabla_{\theta}^{2} R\left(\theta^{*}\right)+\epsilon \cdot \nabla_{\theta}^{2} \mathcal{L}_{U}\left(u^{\prime}, \theta^{*}\right)\right) \cdot\left(\hat{\theta}-\theta^{*}\right) = 0,
\end{split}
\label{c03_eq_infuse_eq3}
\end{gather}
which gives (please refer to Appendix \ref{appendix_infuse_induce} for details):
\begin{align}
\begin{split}
    \hat{\theta}-\theta^{*} &\approx -\left(\nabla_{\theta}^{2} R\left(\theta^{*}\right)\right)^{-1} \cdot \epsilon \nabla_{\theta} \mathcal{L}_{U}\left(u, \theta\right) \\
    & \doteq -\epsilon \cdot H_{\theta}^{-1} \nabla_{\theta} \mathcal{L}_{U}\left(u, \theta\right).
\end{split}
\label{c03_eq_infuse_eq4}
\end{align}

With the help of the chain rule $\frac{\mathrm{d} \mathcal{L}}{\mathrm{d} \epsilon} = \frac{\mathrm{d} \mathcal{L}}{\mathrm{d} \theta} \cdot \frac{\mathrm{d} \theta}{\mathrm{d} \epsilon}$, the importance of an unlabeled example can be estimated:
\begin{align}
\begin{split}
    \text {score}_{\theta}(u)  &=\frac{\mathrm{d} \mathcal{L}\left(V, \theta \right)}{\mathrm{d} \epsilon}
=\nabla_{\theta} \mathcal{L}\left(V, \theta\right)^{\top} \frac{\mathrm{d} \theta}{\mathrm{d} \epsilon} \\
&=-\nabla_{\theta} \mathcal{L}\left(V, \theta\right)^{\top} H_{\theta}^{-1} \nabla_{\theta} \mathcal{L}_{U}\left(u, \theta\right).
\end{split}
\label{c03_eq_infuse_main} 
\end{align}
Equation \ref{c03_eq_infuse_main} is used to compute $\text {score}_{\theta}(u)$ for each unlabeled example. The unlabeled examples with the highest score are preserved to build the core set, and others will be simply dropped. In our implementation, the INFUSE score is calculated batch-wise to reduce the computation overhead. Besides, we use the identity matrix to approximate the inverse Hessian $H_{\theta}^{-1}$ \citep{DBLP:conf/icml/LuketinaRBG16} for efficiency. The last problem is how to compute $\nabla_{\theta} \mathcal{L}\left(V, \theta\right)$ when the ground-truth label of examples in $V$ is unavailable in training. To address this, we propose a feature-level mixup to build a support set $\overline{S}$. Then, the gradient on the validation set is approximated by  $\mathcal{L}\left(\overline{S}, \theta\right)$. Please refer to Appendix \ref{appendix_mixup_feature} for details.

\section{Experiments}
\subsection{Experiment Settings}
We evaluate the effectiveness of our method on standard semi-supervised learning (SSL) datasets: CIFAR-10/100 \citep{krizhevsky2009learning}, SVHN \citep{svhn}, STL-10 \citep{DBLP:journals/jmlr/CoatesNL11}. We follow the commonly used SSL setting \citep{DBLP:conf/nips/SohnBCZZRCKL20} for model training. The keep ratio $k$ controls the size of the core set. For example, with $k=10\%$, the core set size is $10\% \times \left| U \right|$, and the total training steps become $10\%$ of the original iterations.

The model is trained under the most commonly used SSL setting \citep{DBLP:conf/nips/SohnBCZZRCKL20}. The total number of iterations is $2^{20}$ (segmented into 1024 epochs) and batch-size of labeled/unlabeled data is 64/448. We use SGD to optimize the parameters. The learning rate is initially set as $\eta_0=0.03$ with a cosine learning rate decay schedule as $\eta = \eta_0 \cos{(\frac{7 \pi k}{16K})}$, where $k$ is the current iteration and $K$ is the total iterations. 

As for VCC, the size of random noise $z$ is set as 16 for best performance. To reduce the computation overhead, the encoder $q_\phi$ and decoder $p_\theta$ are MLPs with 2 hidden layers (with dimensions 256 and 64). $\lambda_{VCC}$ is set as 2.0.

In INFUSE, the core set is updated for every 40 epochs, and the total number of iterations is adjusted with the keep ratio $k$. Take $k=10\%$ for example, the amount of examples in core set is $10\% \times \left| U \right|$ and the total steps is $10\% \times 2^{20}$. 

\subsection{Main Results}
In this section, we present the effectiveness of VCC and INFUSE individually and then combine them to achieve more efficient and accurate pseudo label selection in SSL.

As mentioned earlier, VCC is a general confidence calibration plugin, allowing flexible combinations with existing SSL methods. In our experiments, we choose popular methods like FixMatch \citep{DBLP:conf/nips/SohnBCZZRCKL20}, FlexMatch \citep{DBLP:conf/nips/ZhangWHWWOS21}, and SimMatch \citep{DBLP:conf/cvpr/ZhengYHWQX22} as the basic modules to build VCC-FixMatch, VCC-FlexMatch, and VCC-SimMatch. The reported values are the mean and standard deviation of three independent trials for each setting, as shown in Table \ref{tab_vcc_main}. All three baseline methods (FixMatch, FlexMatch, SimMatch) exhibit accuracy improvements when combined with VCC for confidence calibration. Particularly, the improvements with VCC are more pronounced when the amount of labeled examples is small. For instance, on CIFAR-100 with only 400 labeled examples, VCC-FlexMatch reduces the error rate from 46.47\% to 43.31\% (-3.16\%). A similar boost is observed on the STL-10 dataset, as shown in Table \ref{stl10_results}, where VCC reduces the error rate of FixMatch by 5.34\% (from 35.97\% to 30.63\%) with only 40 labels.

To further understand the source of the accuracy improvement with VCC, we compute the calibration error of different methods. As shown in Table \ref{appendix_sup_vcc_calibration_results}, both VCC-FixMatch and VCC-FlexMatch achieve lower calibration errors compared to the baseline methods across various settings. VCC-SimMatch also achieves lower Expected Calibration Error (ECE) and Average Calibration Error (ACE) metrics when only 400 labeled examples are available. However, the Maximum Calibration Error (MCE) metric deteriorates, attributed to MCE considering the worst-calibrated bucket and introducing some fluctuations. Under the setting of using 10,000 labeled examples, the results of VCC-SimMatch and SimMatch are very close. This is partly because a larger number of labeled examples can naturally improve the model's performance and reduce the calibration error. Additionally, SimMatch uses instance similarity for rescaling the confidence score, which may reduce the benefits brought by VCC.

The results of INFUSE and other core set selection methods (e.g., RETRIEVE \citep{DBLP:conf/nips/KillamsettyZCI21}) are shown in Table \ref{tab_infuse_main}. On CIFAR-10 dataset, INFUSE achieves a lower error rate (6.29\%) using only 10\% of the examples, indicating the redundancy of original unlabeled data and underscoring the significance of core set selection in SSL. With increasing keep ratio, the gap between INFUSE and the non-pruned setting becomes smaller. For example, on the CIFAR-100 dataset with 2500 labeled data and a keep ratio of 40\%, INFUSE achieves an error rate of 26.47\%, while the baseline is 26.49\%. Compared with other core set selection methods, INFUSE also achieves lower error rates in most settings.

The results demonstrate the effectiveness of VCC and INFUSE individually. By combining them, we propose the VCC-INFUSE method, with results shown in Table \ref{tab_vcc_with_infuse}. VCC-INFUSE achieves a better trade-off between model performance and computation costs. Compared to FlexMatch, VCC-INFUSE-FlexMatch not only reduces the error rate from 26.49\% to 25.41\% but also decreases the training time from 223.96 GPU Hours to 115.47 GPU Hours (-48.44\%).

\begin{table*}[t]
\centering
\caption{Comparison of error rate (\%) for core set selection on different datasets and example keep raito (from 10\% to 60\%).}
\scalebox{0.85}{\begin{tabular}{c|ccc|cc|ccc|cc|cc|cc}
\toprule
\multirow{3}{*}{\textbf{Method}} & \multicolumn{5}{c|}{\textbf{CIFAR-10}} & \multicolumn{5}{c|}{\textbf{CIFAR-100}} & \multicolumn{2}{c|}{\textbf{STL-10}} & \multicolumn{2}{c}{\textbf{SVHN}} \\
\cmidrule(lr){2-6} \cmidrule(lr){7-11} \cmidrule(lr){12-13}  \cmidrule(lr){14-15}
& \multicolumn{3}{c|}{\textbf{250 label}} & \multicolumn{2}{c|}{\textbf{4000 label}} & \multicolumn{3}{c|}{\textbf{2500 label}} & \multicolumn{2}{c|}{\textbf{10000 label}} & \multicolumn{2}{c}{\textbf{250 label}} & \multicolumn{2}{c}{\textbf{250 label}} \\
\cmidrule(lr){2-4} \cmidrule(lr){5-6} \cmidrule(lr){7-9} \cmidrule(lr){10-11} \cmidrule(lr){12-13} \cmidrule(lr){14-15}
& $10\%$ & $20\%$ & $40\%$ & $40\%$ & $60\%$ & $10\%$ & $20\%$ & $40\%$ & $40\%$ & $60\%$ & $10\%$ & $20\%$ & $10\%$ & $20\%$ \\
\midrule
Random & 9.12 & 6.87 & 6.51 & 5.26 & 5.01 & 31.55 & 31.11 & 28.86 & 23.19 & 22.51 & 16.62 & 14.37 & 3.85 & 4.65  \\
Earlystop & 7.47 & 6.03 & 6.85 & 4.86 & 4.52 & 29.21 & 28.85 & 27.30 & 23.03 & 22.61  & 16.31 & 13.20 & 2.93 & 3.08  \\
EL2N & 8.55 & 7.47 & 6.70 & 4.94 & 4.54 & 31.55 & 31.27 & 28.42 & 23.12 & 22.21  & 16.27 & 12.92 & 3.66 & 3.61 \\ 
GradMatch & 6.71 & 5.87 & 5.60 & 4.72 & 4.45 & 28.95 & 28.48 & 26.71 & 22.72 & 22.21 & 16.05 & 12.90 &  2.90 & 2.63 \\
RETRIEVE & 6.60 & 6.02 & 5.48 & 4.68 & 4.41 & \textbf{28.75} & 28.34 & 26.68 & 22.56 & 22.18 & 16.05 & 12.90 &  2.90 & 2.63 \\
INFUSE (Ours) & \textbf{6.29} & \textbf{5.69} & \textbf{5.33} & \textbf{4.51} & \textbf{4.34} & 28.83 & \textbf{28.05} & \textbf{26.47} & \textbf{22.28} & \textbf{21.97} & \textbf{15.84} & \textbf{12.71} & \textbf{2.61} & \textbf{2.46} \\
\hline
Full Unlabeled Data & \multicolumn{3}{c|}{4.98} & \multicolumn{2}{c|}{4.19} & \multicolumn{3}{c|}{26.49} & \multicolumn{2}{c|}{21.90} & \multicolumn{2}{c|}{8.23} & \multicolumn{2}{c}{3.80} \\
\bottomrule
\end{tabular}}%
\label{tab_infuse_main}%
\end{table*}%

\begin{table}[!t]
\centering
\caption{The error rate and training time of different methods on CIFAR-100 dataset with 2500 labeled data. The GPU Hours metric is calculated based on the A100 GPU.}
\scalebox{0.8}{
\begin{tabular}{c|c|c}
\toprule
Method & Error Rate (\%) & Training time (GPU Hours) \\
\midrule
Dash(RandAug) & 27.15 & - \\
MPL & 27.71 & - \\
FixMatch & 28.03 & 221.91 \\ 
FlexMatch & 26.49 & 223.96 \\ \hline
VCC-FlexMatch (Ours) & \textbf{\textcolor{blue}{25.26}} & \textcolor{blue}{253.53} \\ \hline
\makecell{VCC-INFUSE-FlexMatch \\ (Ours, keep raito=40\%)} & \textcolor{red}{25.41} & \textbf{\textcolor{red}{115.47}} \\
\bottomrule
\end{tabular}%
}
\label{tab_vcc_with_infuse}%
\end{table}%

\begin{table}[!h]
\centering
\caption{The error rate of VCC and other calibration methods on CIFAR-100 dataset with 2500 labeled examples.}
\label{appendix_sup_vcc_vs_calibration}
\scalebox{0.7}{\begin{tabular}{c|cccc}
\toprule
Method & ER(\%) & ECE & MCE & ACE \\ \hline
FlexMatch & 26.49 & 0.169 & 0.369 & 0.173 \\ \hline
FlexMatch + Ensemble-TS \citep{DBLP:conf/icml/ZhangKH20} & 26.36 & 0.165 & 0.382 & 0.174 \\
FlexMatch + MMCE \citep{DBLP:conf/icml/KumarSJ18} & 28.44 & 0.182 & 0.374 & 0.185 \\
VCC-FlexMatch (Ours) & \textbf{25.26} & \textbf{0.147} & \textbf{0.324} & \textbf{0.163} \\ \bottomrule
\end{tabular}}%
\end{table}

\begin{table}[!ht]
\centering
\caption{The Error Rate (ER) and calibration errors of VCC when different consistency score is disabled while approximating the $\tilde{r}_u$. Tested under CIFAR-100 dataset with 2500 labeled examples.}
\label{c04_tab_vcc_score_ablation}
\scalebox{0.7}{
\begin{tabular}{ccc|c|c|c|c}
\hline
\textbf{ensemble socre} & \textbf{temporal score} & \textbf{view score} & \textbf{ER(\%)} & \textbf{ECE} & \textbf{MCE} & \textbf{ACE} \\ \hline
\checkmark & \checkmark & \ding{55} & 25.45 & 0.148 & 0.328 & 0.167 \\
\checkmark & \ding{55} & \checkmark & 25.97 & 0.166 & 0.352 & 0.168 \\
\ding{55} & \checkmark & \checkmark & 25.65 & 0.153 & 0.337 & 0.169 \\ \hline
\checkmark & \checkmark & \checkmark & \textbf{25.26} & \textbf{0.147} & \textbf{0.324} & \textbf{0.163} \\
\bottomrule
\end{tabular}}
\end{table}

\subsection{Ablation Study}
We utilize view consistency, temporal consistency, and ensemble consistency for estimating $\tilde{r}$. These three consistency scores are designed to reflect the stability of predictions from different perspectives. To analyze their contributions, we conduct an ablation study, and the result is shown in Table \ref{c04_tab_vcc_score_ablation}. As observed, each consistency score contributes to the estimation of a more accurate $\tilde{r}$, resulting in a lower error rate.

\subsection{Effectiveness of VCC}
In VCC, we initially approximate calibrated confidence to obtain $\tilde{r}_u$. Subsequently, we use a Variational Autoencoder (VAE) to reconstruct it, yielding $r_u$, which is employed in pseudo label selection. The objective of reconstruction is to mitigate the randomness associated with statistical approximation. To demonstrate its necessity, we conduct an ablation study. As shown in Table \ref{appendix_sup_vcc_recon}, VCC with reconstruction further reduces the error rate by 0.50\%.

\begin{table}[!t]
\centering
\caption{Error rates of VCC with or without reconstructing calibrated confidence, on CIFAR-100 with 2500 labeled examples.}
\label{appendix_sup_vcc_recon}
\scalebox{0.8}{
\begin{tabular}{c|c|c|c|c}
\hline
\textbf{Reconstruct $\tilde{r}_u$ by VAE} & \textbf{Error Rate (\%)} & \textbf{ECE} & \textbf{MCE} & \textbf{ACE} \\ \hline
 \ding{55} & 25.76  & 0.160 & 0.411 & 0.168  \\
 \checkmark & \textbf{25.26} & \textbf{0.147} & \textbf{0.324} & \textbf{0.163} \\
\bottomrule
\end{tabular}}
\end{table}

\subsection{VCC vs. Other Calibration Methods}
While most calibration methods designed for fully-supervised settings may not be directly suitable for SSL, pseudo labels can be used to approximate ground truth. We choose Ensemble-TS \citep{DBLP:conf/icml/ZhangKH20} and MMCE \citep{DBLP:conf/icml/KumarSJ18} as baselines to compare with VCC. As depicted in Table \ref{appendix_sup_vcc_vs_calibration}, MMCE exhibits the highest error rate (28.44\%). The reason is that MMCE directly employs pseudo labels to calculate calibration regularization, which may introduce noise due to incorrect pseudo labels. In contrast, Ensemble-TS, using pseudo labels to search for optimal parameter scaling, alleviates the issue to some extent (ER=26.36\%). In comparison, VCC achieves the lowest error rate of 25.26\% and the best calibration performance. 


\section{Conclusion}
In this study, we addressed the challenges associated with leveraging large-scale unlabeled data in SSL and proposed two novel methods, VCC and INFUSE, to enhance the effectiveness and efficiency of data selection. As a versatile plugin, VCC significantly improves the accuracy of FixMatch, FlexMatch, and SimMatch across multiple datasets. Simultaneously, INFUSE achieves competitive or even lower error rates with partial unlabeled data. By combining the two methods, VCC-INFUSE achieves a lower error rate with less computational overhead. Future work will involve extending VCC-INFUSE to various SSL tasks, such as object detection and segmentation, to assess its generalization.

\section*{Acknowledgements}
This work was supported by National Key R\&D Program of China (No.2018AAA0100300) and the funding of China Tower.

\clearpage
\bibliographystyle{named}
\bibliography{ijcai24}

\begin{thebibliography}{}

\bibitem[\protect\citeauthoryear{Berthelot \bgroup \em et al.\egroup }{2019}]{DBLP:conf/nips/BerthelotCGPOR19}
David Berthelot, Nicholas Carlini, Ian~J. Goodfellow, Nicolas Papernot, Avital Oliver, and Colin Raffel.
\newblock {MixMatch: {A} Holistic Approach to Semi-Supervised Learning}.
\newblock In {\em Advances in Neural Information Processing Systems, Vancouver, British Columbia, Canada}, 2019.

\bibitem[\protect\citeauthoryear{Berthelot \bgroup \em et al.\egroup }{2020}]{DBLP:conf/iclr/BerthelotCCKSZR20}
David Berthelot, Nicholas Carlini, Ekin~D. Cubuk, Alex Kurakin, Kihyuk Sohn, Han Zhang, and Colin Raffel.
\newblock {ReMixMatch: Semi-Supervised Learning with Distribution Matching and Augmentation Anchoring}.
\newblock In {\em International Conference on Learning Representations, Addis Ababa, Ethiopia}, 2020.

\bibitem[\protect\citeauthoryear{Chen \bgroup \em et al.\egroup }{2023}]{DBLP:conf/iclr/0102TFW0S0RS23}
Hao Chen, Ran Tao, Yue Fan, Yidong Wang, Jindong Wang, Bernt Schiele, Xing Xie, Bhiksha Raj, and Marios Savvides.
\newblock Softmatch: Addressing the quantity-quality tradeoff in semi-supervised learning.
\newblock In {\em The Eleventh International Conference on Learning Representations, {ICLR} 2023, Kigali, Rwanda, May 1-5, 2023}, 2023.

\bibitem[\protect\citeauthoryear{Coates \bgroup \em et al.\egroup }{2011}]{DBLP:journals/jmlr/CoatesNL11}
Adam Coates, Andrew~Y. Ng, and Honglak Lee.
\newblock {An Analysis of Single-Layer Networks in Unsupervised Feature Learning}.
\newblock In {\em Proceedings of the Fourteenth International Conference on Artificial Intelligence and Statistics, Fort Lauderdale, USA}, 2011.

\bibitem[\protect\citeauthoryear{Deng \bgroup \em et al.\egroup }{2009}]{DBLP:conf/cvpr/DengDSLL009}
Jia Deng, Wei Dong, Richard Socher, Li{-}Jia Li, Kai Li, and Li~Fei{-}Fei.
\newblock {ImageNet: {A} Large-Scale Hierarchical Image Database}.
\newblock In {\em Conference on Computer Vision and Pattern Recognition, Miami, Florida, {USA}}, 2009.

\bibitem[\protect\citeauthoryear{Feng \bgroup \em et al.\egroup }{2024}]{DBLP:conf/aaai/FengXF024}
Qianhan Feng, Lujing Xie, Shijie Fang, and Tong Lin.
\newblock Bacon: Boosting imbalanced semi-supervised learning via balanced feature-level contrastive learning.
\newblock In {\em Thirty-Eighth {AAAI} Conference on Artificial Intelligence, {AAAI} 2024}, 2024.

\bibitem[\protect\citeauthoryear{Gal and Ghahramani}{2016}]{DBLP:conf/icml/GalG16}
Yarin Gal and Zoubin Ghahramani.
\newblock {Dropout as a Bayesian Approximation: Representing Model Uncertainty in Deep Learning}.
\newblock In {\em Proceedings of the 33nd International Conference on Machine Learning, New York City, NY, USA}, 2016.

\bibitem[\protect\citeauthoryear{Ganev and Aitchison}{2020}]{ganev2020semi}
Stoil Ganev and Laurence Aitchison.
\newblock {Semi-supervised Learning Objectives as Log-likelihoods in a Generative Model of Data Curation}.
\newblock {\em arXiv preprint arXiv:2008.05913}, 2020.

\bibitem[\protect\citeauthoryear{Guo and Li}{2022}]{DBLP:conf/icml/GuoL22}
Lan{-}Zhe Guo and Yu{-}Feng Li.
\newblock {Class-Imbalanced Semi-Supervised Learning with Adaptive Thresholding}.
\newblock In {\em International Conference on Machine Learning, Baltimore, Maryland, {USA}}, 2022.

\bibitem[\protect\citeauthoryear{Guo \bgroup \em et al.\egroup }{2017}]{DBLP:conf/icml/GuoPSW17}
Chuan Guo, Geoff Pleiss, Yu~Sun, and Kilian~Q. Weinberger.
\newblock {On Calibration of Modern Neural Networks}.
\newblock In {\em International Conference on Machine Learning, Sydney, NSW, Australia}, 2017.

\bibitem[\protect\citeauthoryear{Killamsetty \bgroup \em et al.\egroup }{2021a}]{DBLP:conf/icml/KillamsettySRDI21}
KrishnaTeja Killamsetty, Durga Sivasubramanian, Ganesh Ramakrishnan, Abir De, and Rishabh~K. Iyer.
\newblock {{GRAD-MATCH:} Gradient Matching based Data Subset Selection for Efficient Deep Model Training}.
\newblock In {\em Proceedings of the International Conference on Machine Learning, Virtual Event}, 2021.

\bibitem[\protect\citeauthoryear{Killamsetty \bgroup \em et al.\egroup }{2021b}]{DBLP:conf/nips/KillamsettyZCI21}
KrishnaTeja Killamsetty, Xujiang Zhao, Feng Chen, and Rishabh~K. Iyer.
\newblock {{RETRIEVE:} Coreset Selection for Efficient and Robust Semi-Supervised Learning}.
\newblock In {\em Advances in Neural Information Processing Systems, virtual}, 2021.

\bibitem[\protect\citeauthoryear{Kim \bgroup \em et al.\egroup }{2020}]{DBLP:conf/nips/KimHPYHS20}
Jaehyung Kim, Youngbum Hur, Sejun Park, Eunho Yang, Sung~Ju Hwang, and Jinwoo Shin.
\newblock {Distribution Aligning Refinery of Pseudo-label for Imbalanced Semi-supervised Learning}.
\newblock In {\em Advances in Neural Information Processing Systems, virtual}, 2020.

\bibitem[\protect\citeauthoryear{Kingma and Welling}{2014}]{DBLP:journals/corr/KingmaW13}
Diederik~P. Kingma and Max Welling.
\newblock {Auto-Encoding Variational Bayes}.
\newblock In {\em International Conference on Learning Representations, Banff, AB, Canada}, 2014.

\bibitem[\protect\citeauthoryear{Koh and Liang}{2017}]{DBLP:conf/icml/KohL17}
Pang~Wei Koh and Percy Liang.
\newblock Understanding black-box predictions via influence functions.
\newblock In {\em Proceedings of the 34th International Conference on Machine Learning, {ICML} 2017, Sydney, NSW, Australia, 6-11 August 2017}, 2017.

\bibitem[\protect\citeauthoryear{Krizhevsky \bgroup \em et al.\egroup }{2009}]{krizhevsky2009learning}
Alex Krizhevsky, Geoffrey Hinton, et~al.
\newblock {Learning Multiple Layers of Features from Tiny Images}.
\newblock In {\em Doctoral dissertation, University of Toronto}, 2009.

\bibitem[\protect\citeauthoryear{Kumar \bgroup \em et al.\egroup }{2018}]{DBLP:conf/icml/KumarSJ18}
Aviral Kumar, Sunita Sarawagi, and Ujjwal Jain.
\newblock {Trainable Calibration Measures For Neural Networks From Kernel Mean Embeddings}.
\newblock In {\em International Conference on Machine Learning, Stockholm, Sweden}, 2018.

\bibitem[\protect\citeauthoryear{Laine and Aila}{2016}]{DBLP:journals/corr/LaineA16}
Samuli Laine and Timo Aila.
\newblock {Temporal Ensembling for Semi-Supervised Learning}.
\newblock {\em arXiv preprint arXiv:1610.02242}, 2016.

\bibitem[\protect\citeauthoryear{Lee and others}{2013}]{lee2013pseudo}
Dong-Hyun Lee et~al.
\newblock Pseudo-label: The simple and efficient semi-supervised learning method for deep neural networks.
\newblock In {\em Workshop on challenges in representation learning, ICML}, 2013.

\bibitem[\protect\citeauthoryear{Lee \bgroup \em et al.\egroup }{2021}]{lee2021abc}
Hyuck Lee, Seungjae Shin, and Heeyoung Kim.
\newblock {ABC}: Auxiliary balanced classifier for class-imbalanced semi-supervised learning.
\newblock In {\em Advances in Neural Information Processing Systems}, 2021.

\bibitem[\protect\citeauthoryear{Li \bgroup \em et al.\egroup }{2021}]{DBLP:conf/iccv/0001XH21}
Junnan Li, Caiming Xiong, and Steven C.~H. Hoi.
\newblock Comatch: Semi-supervised learning with contrastive graph regularization.
\newblock In {\em {IEEE/CVF} International Conference on Computer Vision, Montreal, QC, Canada, October 10-17}, 2021.

\bibitem[\protect\citeauthoryear{Lin \bgroup \em et al.\egroup }{2014}]{DBLP:conf/eccv/LinMBHPRDZ14}
Tsung{-}Yi Lin, Michael Maire, Serge~J. Belongie, James Hays, Pietro Perona, Deva Ramanan, Piotr Doll{\'{a}}r, and C.~Lawrence Zitnick.
\newblock {Microsoft {COCO:} Common Objects in Context}.
\newblock In {\em European Conference on Computer Vision, Zurich, Switzerland}, 2014.

\bibitem[\protect\citeauthoryear{Luketina \bgroup \em et al.\egroup }{2016}]{DBLP:conf/icml/LuketinaRBG16}
Jelena Luketina, Tapani Raiko, Mathias Berglund, and Klaus Greff.
\newblock {Scalable Gradient-Based Tuning of Continuous Regularization Hyperparameters}.
\newblock In {\em International Conference on Machine Learning, New York City, NY, USA}, 2016.

\bibitem[\protect\citeauthoryear{Miyato \bgroup \em et al.\egroup }{2019}]{DBLP:journals/pami/MiyatoMKI19}
Takeru Miyato, Shin{-}ichi Maeda, Masanori Koyama, and Shin Ishii.
\newblock {Virtual Adversarial Training: {A} Regularization Method for Supervised and Semi-Supervised Learning}.
\newblock {\em {IEEE Transactions on Pattern Analysis and Machine Intelligence}}, 41(8):1979--1993, 2019.

\bibitem[\protect\citeauthoryear{Netzer \bgroup \em et al.\egroup }{2011}]{svhn}
Yuval Netzer, Tao Wang, Adam Coates, Alessandro Bissacco, Bo~Wu, and Andrew~Y. Ng.
\newblock {Reading Digits in Natural Images with Unsupervised Feature Learning}.
\newblock In {\em NIPS Workshop on Deep Learning and Unsupervised Feature Learning}, 2011.

\bibitem[\protect\citeauthoryear{Nixon \bgroup \em et al.\egroup }{2019}]{DBLP:conf/cvpr/NixonDZJT19}
Jeremy Nixon, Michael~W. Dusenberry, Linchuan Zhang, Ghassen Jerfel, and Dustin Tran.
\newblock {Measuring Calibration in Deep Learning}.
\newblock In {\em {IEEE} Conference on Computer Vision and Pattern Recognition Workshops, Long Beach, CA, USA}, 2019.

\bibitem[\protect\citeauthoryear{Oh \bgroup \em et al.\egroup }{2022}]{DBLP:conf/cvpr/Oh0K22}
Youngtaek Oh, Dong{-}Jin Kim, and In~So Kweon.
\newblock {{DASO:} Distribution-Aware Semantics-Oriented Pseudo-label for Imbalanced Semi-Supervised Learning}.
\newblock In {\em {IEEE/CVF} Conference on Computer Vision and Pattern Recognition, New Orleans, LA, USA}, 2022.

\bibitem[\protect\citeauthoryear{Paul \bgroup \em et al.\egroup }{2021}]{DBLP:conf/nips/PaulGD21}
Mansheej Paul, Surya Ganguli, and Gintare~Karolina Dziugaite.
\newblock {Deep Learning on a Data Diet: Finding Important Examples Early in Training}.
\newblock In {\em Advances in Neural Information Processing Systems, virtual}, 2021.

\bibitem[\protect\citeauthoryear{Pham \bgroup \em et al.\egroup }{2021}]{DBLP:conf/cvpr/PhamDXL21}
Hieu Pham, Zihang Dai, Qizhe Xie, and Quoc~V. Le.
\newblock {Meta Pseudo Labels}.
\newblock In {\em {IEEE} Conference on Computer Vision and Pattern Recognition, virtual}, 2021.

\bibitem[\protect\citeauthoryear{Sohn \bgroup \em et al.\egroup }{2020}]{DBLP:conf/nips/SohnBCZZRCKL20}
Kihyuk Sohn, David Berthelot, Nicholas Carlini, Zizhao Zhang, Han Zhang, Colin Raffel, Ekin~Dogus Cubuk, Alexey Kurakin, and Chun{-}Liang Li.
\newblock {FixMatch: Simplifying Semi-Supervised Learning with Consistency and Confidence}.
\newblock In {\em Advances in Neural Information Processing Systems, Virtual}, 2020.

\bibitem[\protect\citeauthoryear{Tarvainen and Valpola}{2017}]{DBLP:conf/nips/TarvainenV17}
Antti Tarvainen and Harri Valpola.
\newblock {Mean Teachers are Better Role Models: Weight-Averaged Consistency Targets Improve Semi-Supervised Deep Learning Results}.
\newblock In {\em Advances in Neural Information Processing Systems, Toulon, France}, 2017.

\bibitem[\protect\citeauthoryear{Wang \bgroup \em et al.\egroup }{2022}]{DBLP:conf/cvpr/WangWLY22}
Xudong Wang, Zhirong Wu, Long Lian, and Stella~X. Yu.
\newblock {Debiased Learning from Naturally Imbalanced Pseudo-Labels}.
\newblock In {\em {IEEE/CVF} Conference on Computer Vision and Pattern Recognition, New Orleans, LA, USA}, 2022.

\bibitem[\protect\citeauthoryear{Xie \bgroup \em et al.\egroup }{2020}]{DBLP:conf/nips/XieDHL020}
Qizhe Xie, Zihang Dai, Eduard~H. Hovy, Thang Luong, and Quoc Le.
\newblock {Unsupervised Data Augmentation for Consistency Training}.
\newblock In {\em Advances in Neural Information Processing Systems, Virtual}, 2020.

\bibitem[\protect\citeauthoryear{Xu \bgroup \em et al.\egroup }{2015}]{DBLP:journals/tip/XuT015}
Chang Xu, Dacheng Tao, and Chao Xu.
\newblock {Multi-View Learning With Incomplete Views}.
\newblock {\em {IEEE} Trans. Image Process.}, 24(12):5812--5825, 2015.

\bibitem[\protect\citeauthoryear{Xu \bgroup \em et al.\egroup }{2021}]{DBLP:conf/icml/XuSYQLSLJ21}
Yi~Xu, Lei Shang, Jinxing Ye, Qi~Qian, Yu{-}Feng Li, Baigui Sun, Hao Li, and Rong Jin.
\newblock {Dash: Semi-Supervised Learning with Dynamic Thresholding}.
\newblock In {\em Proceedings of the International Conference on Machine Learning, Virtual Event}, 2021.

\bibitem[\protect\citeauthoryear{Zhang \bgroup \em et al.\egroup }{2020}]{DBLP:conf/icml/ZhangKH20}
Jize Zhang, Bhavya Kailkhura, and Thomas~Yong{-}Jin Han.
\newblock {Mix-N-Match : Ensemble and Compositional Methods for Uncertainty Calibration in Deep Learning}.
\newblock In {\em International Conference on Machine Learning, Virtual}, 2020.

\bibitem[\protect\citeauthoryear{Zhang \bgroup \em et al.\egroup }{2021}]{DBLP:conf/nips/ZhangWHWWOS21}
Bowen Zhang, Yidong Wang, Wenxin Hou, Hao Wu, Jindong Wang, Manabu Okumura, and Takahiro Shinozaki.
\newblock {FlexMatch: Boosting Semi-Supervised Learning with Curriculum Pseudo Labeling}.
\newblock In {\em Advances in Neural Information Processing, Virtual}, 2021.

\bibitem[\protect\citeauthoryear{Zheng \bgroup \em et al.\egroup }{2022}]{DBLP:conf/cvpr/ZhengYHWQX22}
Mingkai Zheng, Shan You, Lang Huang, Fei Wang, Chen Qian, and Chang Xu.
\newblock {SimMatch: Semi-supervised Learning with Similarity Matching}.
\newblock In {\em Conference on Computer Vision and Pattern Recognition, New Orleans, LA, USA}, 2022.

\bibitem[\protect\citeauthoryear{Zhou \bgroup \em et al.\egroup }{2020}]{DBLP:conf/icml/ZhouWB20}
Tianyi Zhou, Shengjie Wang, and Jeff~A. Bilmes.
\newblock {Time-Consistent Self-Supervision for Semi-Supervised Learning}.
\newblock In {\em International Conference on Machine Learning, Virtual}, 2020.

\end{thebibliography}

\clearpage
\appendix
\section*{Appendix}
\section{Optimizing VAE in VCC}
\label{appendix_vcc_formulation}
In Equation \ref{vc_eq2}, we use another distribution $q_\phi(z|c,x)$ as the approximation of $p_\theta(z)$:

\begin{equation}
\begin{split}
& \log p_\theta(r |c,x)\\
& =\int_z q_\phi(z|c,x) \log p_\theta(r|c,x) \mathrm{d}z\\
& =\int_z q_\phi(z|c,x) \log \frac{p_\theta(r|c,z,x)p_\theta(z|c,x)}{p_\theta(z|r,c,x)} \mathrm{d}z\\
& = \int_z q_\phi(z|c,x) \log \left( \frac{p_\theta(r|c,z,x)p_\theta(z|c,x)}{p_\theta(z|r,c,x)} \frac{q_\phi(z|c,x)}{q_\phi(z|c,x)} \right) \mathrm{d}z\\
& = \int_z q_\phi(z|c,x) ( \log \frac{p_\theta(r|c,z,x)p_\theta(z|c,x)}{q_\phi(z|c,x)}\\
& \qquad\qquad\qquad\qquad\qquad\qquad\qquad + \log \frac{q_\phi(z|c,x)}{p_\theta(z|r,c,x)} ) \mathrm{d}z\\
& = \int_z q_\phi(z|c,x) \log \frac{p_\theta(r|c,z,x)p_\theta(z|c,x)}{q_\phi(z|c,x)} \mathrm{d}z \\
& \qquad\qquad\qquad\qquad\qquad + D_{KL}(q_\phi(z|c,x) \Vert p_\theta(z|c,r,x))\\
& \ge \int_z q_\phi(z|c,x) \log \frac{p_\theta(r|c,z,x)p_\theta(z|c,x)}{q_\phi(z|c,x)} \mathrm{d}z,
\end{split}
\end{equation}
where the second equation employs the Bayes' theorem: $p(r)=p(r,z)/p(z|r)=p(r|z)p(z)/p(z|r)$.
In Inequality \ref{vc_eq2}, the non-negative Kullback-Leibler divergence $D_{KL}(q \Vert p)$ cannot be directly computed and the remaining part is called the Evidence Lower Bound (ELBO) of variational. 
To find the optimal $q_\phi(z|c,x)$ to approximate $p_\theta(z|c,r,x)$, the ELBO requires to be maximized. The Inequality \ref{vc_eq2} can be further rewritten as:
\begin{equation}
    \begin{split}
        &\log p_\theta(r|c,x) \ge \int_z q_\phi(z|c,x) \log \frac{p_\theta(r|c,z,x)p_\theta(z|c,x)}{q_\phi(z|c,x)} \mathrm{d}z \\
    &= \mathbb{E}_{q_\phi (z|c,x)} \log p_\theta (r|c,z,x) - 
    D_{KL} (q_\phi (z|c,x) \Vert p_\theta(z|c,x)).
    \end{split}
\end{equation}
The first term is the likelihood of calibration reconstruction, where $q_\phi(z|c,x)$ is the encoder to infer the hidden variable $z$, and $p_\theta(r|c,z,x)$ is the decoder to recover a calibrated confidence $r$.
Under the Gaussian assumptions of the prior $p_\theta(z|c,x) \sim \mathcal{N}(0,1)$ and the approximator $q_\phi(z|c,x) \sim \mathcal{N}(\mu(c,x), \sigma^2(c,x))$, the second term is equal to:
\begin{equation}
    \begin{split}
    & D_{KL} (q_\phi (z|c,x) \Vert p_\theta(z|c,x)) \\
    &= \int_z q_\phi (z|c,x) (\log q_\phi (z|c,x) - \log p_\theta (z|c,x)) \mathrm{d}z \\
    &= \int_z q_\phi (z|c,x) (\log \frac{1}{\sqrt{2 \pi \sigma^2}} e^{-\frac{(z-\mu)^2}{2\sigma^2}} - \log \frac{1}{\sqrt{2 \pi}} e^{-\frac{z^2}{2}} ) \mathrm{d}z \\
    &= \int_z q_\phi (z|c,x) \Big(-\frac{(z-\mu)^2}{2\sigma^2} + \log \frac{1}{\sqrt{2 \pi \sigma^2}} + \frac{z^2}{2} \\ 
    & \qquad\qquad\qquad\qquad\qquad\qquad\qquad\qquad\qquad- \log \frac{1}{\sqrt{2 \pi}}  \Big) \mathrm{d}z  \notag
    \end{split}
\end{equation}
\begin{equation}
    \begin{split}
    &= -\int_z q_\phi (z|c,x) \log {\sigma} \mathrm{d}z + \int_z q_\phi (z|c,x) \frac{z^2}{2} \mathrm{d}z \\ 
    & \qquad\qquad\qquad\qquad\qquad\quad- \int_z q_\phi (z|c,x) \frac{(z-\mu)^2}{2\sigma^2} \mathrm{d}z  \\
    &= - \log {\sigma} + \mathbb{E}_{z\sim q_\phi} \left[ \frac{z^2}{2} \right] - \mathbb{E}_{z\sim q_\phi} \left[ \frac{(z-\mu)^2}{2\sigma^2} \right] \\
    &= - \log {\sigma} + \frac{1}{2} \big( (\mathbb{E}_{z\sim q_\phi} [z])^2  + Var(z) \big) - \frac{1}{2\sigma^2} Var(z) \\
    &= - \log {\sigma} + \frac{\mu^2 + \sigma^2}{2} - \frac{1}{2},
    \end{split}
\end{equation}
where the second last equation employs the variance lemma: $\mathbb{E}\left[z^2\right] = (\mathbb{E}\left[z\right])^2 + Var(z)$.
To compute the reconstruction error, the $z$ need to be sampled from $q_\phi(z|c,x)$. We use the reparameterization \cite{DBLP:journals/corr/KingmaW13} trick to address this.
By setting $\epsilon \sim \mathcal{N}(0,1)$, the reparameterization is formulated as
$z = \mu(c,x) + \epsilon \cdot \sigma(c,x)$.

\section{The Detailed Deduction of INFUSE}
\label{appendix_infuse_induce}
To deduce Equation \ref{c03_eq_infuse_eq4} from Equation \ref{c03_eq_infuse_eq3}, we have:
\begin{equation}
    \begin{split}
        \hat{\theta}-\theta^{*}  =- [\nabla_{\theta}^{2} R&\left(\theta^{*}\right)+\epsilon \cdot \nabla_{\theta}^{2} \mathcal{L}_{U}\left(u^{\prime}, \theta^{*}\right)]^{-1} \\ \cdot  
        & \left[\nabla_{\theta} R\left(\theta^{*}\right)+\epsilon \cdot \nabla_{\theta} \mathcal{L}_{U}\left(u^{\prime}, \theta^{*}\right)\right].
    \end{split}
\end{equation}
Here $\nabla_{\theta} R\left(\theta^{*}\right) = 0$, since $\theta^{*}$ is the optimal parameters that minimize $R(\theta)$:
\begin{equation}
    \begin{split}
        \hat{\theta}-\theta^{*} = -[\nabla_{\theta}^{2} R\left(\theta^{*}\right)+\epsilon \cdot \nabla_{\theta}^{2} &\mathcal{L}_{U}\left(u^{\prime}, \theta^{*}\right)]^{-1} \\
        & \cdot \epsilon \nabla_{\theta} \mathcal{L}_{U}\left(u^{\prime}, \theta^{*}\right).
    \end{split}
\end{equation}

Note that $\epsilon$ is an extremely small number. For the entity of $\left[\nabla_{\theta}^{2} R\left(\theta^{*}\right)+\epsilon \cdot \nabla_{\theta}^{2} \mathcal{L}_{U}\left(u^{\prime}, \theta^{*}\right)\right]^{-1}$, the contribution of $\epsilon \cdot \nabla_{\theta}^{2} \mathcal{L}_{U}\left(u^{\prime}, \theta^{*}\right)$ is so small that we can approximately omit it. Now we have:
\begin{align}
\hat{\theta}-\theta^{*} = -\epsilon\left(\nabla_{\theta}^{2} R\left(\theta^{*}\right)\right)^{-1} \nabla_{\theta} \mathcal{L}_{U}\left(u^{\prime}, \theta^{*}\right).
\end{align}
\section{Approximate $\nabla_{\theta} \mathcal{L}\left(V, \theta\right)$ with Mixup}
\label{appendix_mixup_feature}
In Equation \ref{c03_eq_infuse_main}, the first term $\nabla_{\theta} \mathcal{L}\left(V, \theta\right)^{\top}$ is the gradient on the validation set. However, it's infeasible to direct compute since the ground-truth label of examples in $V$ is unavailable in training. One applicable approximation is $\nabla_{\theta} \mathcal{L}\left(V, \theta\right) \approx \nabla_{\theta} \mathcal{L}\left(S, \theta\right)$, i.e. use the gradient on labeled training set. However, the volume of the labeled dataset is small in SSL and the model tends to overfit quickly on that. Hence, $\nabla_{\theta} \mathcal{L}\left(S, \theta\right)$ may bring huge error in practice. Another way is to approximate $\nabla_{\theta} \mathcal{L}\left(V, \theta\right)$ with the gradient of unlabeled dataset. However, the pseudo label can be noisy (especially in the earlier stage of training), which may lead to the wrong gradient.

We argue that the approximation of $\nabla_{\theta} \mathcal{L}\left(V, \theta\right)$ should: (1) be free from the overfitting problem; (2) be calculated with the reliable ground-truth label to ensure the correctness. In this paper, we propose a Mixup-based approximation method. Given the labeled training set $S$, we randomly sample $2K$ examples from it: $\tilde{S}=\left\{(x_i, y_i), i=1\dots 2K\right\}$, followed by the backbone to extract features $h$ for each example: $h_{\tilde{V}}=\left\{ h_i, i=1\dots 2K \right\}$. Then, we apply Mixup to the feature and ground-truth label: $\overline{h_i} = \text{Mixup}\left(h_{2i}, h_{2i+1}\right)$, $\overline{y_i} = \text{Mixup}\left(y_{2i}, y_{2i+1}\right)$
to obtain the support set $\overline{S} = \left\{ (\overline{h_i}, \overline{y_i}), i=1\dots K \right\}$. Finally, the classification head will output the confidence distributions based on the features after Mixup and compute the loss. The gradient $\nabla_{\theta} \mathcal{L}\left(\overline{S}, \theta\right)$ is used as the approximation of $\nabla_{\theta} \mathcal{L}\left({V}, \theta\right)$.

Mixup on labeled examples can provide accurate pseudo labels and alleviate the problem of overfitting. What's more, the feature-level Mixup ensures the input domain is unchanged, so that the backbone network can extract features correctly, making the gradient of support set closer to $\nabla_{\theta} \mathcal{L}\left({V}, \theta\right)$.

\section{Experiments Details}
\label{exp_setting}
As for VCC, we compare it with SimMatch\citep{DBLP:conf/cvpr/ZhengYHWQX22}, FlexMatch\citep{DBLP:conf/nips/ZhangWHWWOS21}, Dash\citep{DBLP:conf/icml/XuSYQLSLJ21}, MPL \citep{DBLP:conf/cvpr/PhamDXL21}, FixMatch\citep{DBLP:conf/nips/SohnBCZZRCKL20}, ReMixMatch\citep{DBLP:conf/iclr/BerthelotCCKSZR20}, UDA\citep{DBLP:conf/nips/XieDHL020}, MixMatch\citep{DBLP:conf/nips/BerthelotCGPOR19}, VAT\citep{DBLP:journals/pami/MiyatoMKI19}, MeanTeacher\citep{DBLP:conf/nips/TarvainenV17} and PL\citep{lee2013pseudo}.
As for core set selection experiments, we use FlexMatch as the SSL method and compare INFUSE with RETRIEVE\citep{DBLP:conf/nips/KillamsettyZCI21}, GradMatch\citep{DBLP:conf/icml/KillamsettySRDI21}, EL2N\citep{DBLP:conf/nips/PaulGD21}, Random\citep{DBLP:conf/icml/KillamsettySRDI21}, and Earlystop\citep{DBLP:conf/icml/KillamsettySRDI21}.

The error rate on test set is used as the main metric to evaluate the effectiveness of our methods. To further study the reduction of calibration error, we introduce ECE/MCE \citep{DBLP:conf/icml/GuoPSW17} and ACE \citep{DBLP:conf/cvpr/NixonDZJT19}.
To calculate the calibration error, the examples in test set into buckets $B_1, B_2, \dots, B_m$ based on confidence scores (for example, samples with confidence scores in the range $[0.90~0.95)$ are assigned to the same bucket).
The Expected Calibration Error (ECE) can be formulated as:
\begin{align}
\text{ECE} = \sum_{i=1}^m \frac{|B_i|}{N} \left| \text{conf}(B_i) - \text{acc}(B_i) \right|,
\label{c04_eq_ece}
\end{align}
where $\text{acc}(B_i)$ and $\text{conf}(B_i)$ represent the average accuracy and average confidence score of examples in bucket $B_i$, respectively.

Unlike ECE, Maximum Calibration Error (MCE) \cite{DBLP:conf/icml/GuoPSW17} measures the model's calibration error in the worst-case scenario. It can be expressed as:
\begin{align}
\text{MCE} = \max\limits_{i} \left| \text{conf}(B_i) - \text{acc}(B_i) \right|,
\end{align}

The boundaries for dividing buckets in ECE and MCE are predefined confidence intervals. On the other hand, Adaptive Calibration Error (ACE) \citep{DBLP:conf/cvpr/NixonDZJT19} aims to ensure an equal number of samples in each bucket during grouping. By ensuring that each bucket contains $\lfloor \frac{n}{m }\rfloor$ samples, the resulting buckets can be denoted as $B^{\prime}_1, B^{\prime}_2, \dots, B^{\prime}_m$. The  formula for ACE is as follows:
\begin{align}
\text{ACE} = \sum_{i=1}^m \frac{|B^{\prime}_i|}{N} \left| \text{conf}(B^{\prime}_i) - \text{acc}(B^{\prime}_i) \right|.
\end{align}

A model with lower ECE/MCE/ACE is expected to be better calibrated.
\section{Experiments Setting of Imbalance SSL}
\label{imbalance_SSL}

In this section, we explore a more difficult but realistic setting: imbalance semi-supervised learning. While the problem of SSL comes from the situation that it's difficult to make labels for all data, collecting balance data for each class could be more challenging since most ground-truth information is not accessible when making dataset. Using imbalance data to train model can make the bias generated during semi-supervised learning become more severe, thus hurting the performance. 

We test the robustness of VCC on this problem to further testify its ability to reduce the bias and produce more accurate pseudo label. We construct a CIFAR-10-based long-tail distribution dataset in which the number of data points exponentially decreases from the largest to the smallest class, i.e., $N_{k}=N_{1}\times \gamma ^{-\frac{k-1}{L-1}}$, where $N_{k}$ stands for the number of labeled data of the k-th class, $\gamma=\frac{N_{1}}{N_{L}}$, and ${L}$ is the total class number. Let $\beta$ represent the ratio of labeled data to all data in the same class. We set $\beta=20\%$, $N_{1}=1000$ and $\gamma=100$ to build the long-tail dataset CIFAR-10-LT. We plug VCC into FixMatch and use the same setting as the main experiment to train the model on CIFAR-10-LT.

To better testify the effectiveness of VCC on this setting, we not only tarin FlexMatch, Fixmatch and SimMatch as comparisions, but also include other Imbalance Semi-supervised Learning methods that focus on exploiting the distribution bias information, such as DASO \citep{DBLP:conf/cvpr/Oh0K22}, DebiasPL \citep{DBLP:conf/cvpr/WangWLY22}, DARP \citep{DBLP:conf/nips/KimHPYHS20} and Adsh \citep{DBLP:conf/icml/GuoL22}. We plug these Imbalance SSL methods into FixMatch for fairness and train them with their own default settings.  The experiments have been given in Table \ref{table:classimbalance}.

\begin{table}[!t]
\centering
\caption{The error rate (\%) of different methods on CIFAR-10-LT under the class imbalance setting.}
\label{table:classimbalance}
{\scalebox{0.85}{
\begin{tabular}{l|c}
\hline
Method & Error Rate (\%) \\ \hline
FixMatch \citep{DBLP:conf/nips/SohnBCZZRCKL20}
& 25.07 \\ \hline
FlexMatch \citep{DBLP:conf/nips/ZhangWHWWOS21} & 25.87 \\ \hline
SimMatch \citep{DBLP:conf/cvpr/ZhengYHWQX22} & 61.54 \\ \hline
FixMatch+DASO \citep{DBLP:conf/cvpr/Oh0K22} & 24.63 \\ \hline
FixMatch+DebiasPL \citep{DBLP:conf/cvpr/WangWLY22} & 24.42 \\ \hline
FixMatch+DARP \citep{DBLP:conf/nips/KimHPYHS20} & 22.93 \\ \hline
FixMatch+Adsh \citep{DBLP:conf/icml/GuoL22} & 21.88 \\ \hline
FixMatch+VCC (Ours)
& \textbf{21.16} \\ \hline
\end{tabular}}
}
\end{table}

\end{document}